# BioFLAIR: Pretrained Pooled Contextualized Embeddings for Biomedical Sequence Labeling Tasks

**Shreyas Sharma**[1]
Manipal Institute of Technology
shreyas.kumar3@learner.manipal.edu

**Ron Daniel, Jr.**
Elsevier Labs
r.daniel@elsevier.com

## Abstract

Biomedical Named Entity Recognition (NER) is a challenging problem in biomedical information processing due to the widespread ambiguity of out of context terms and extensive lexical variations. Performance on bioNER benchmarks continues to improve due to advances like BERT, GPT, and XLNet. FLAIR (1) is an alternative embedding model which is less computationally intensive than the others mentioned. We test FLAIR and its pretrained PubMed embeddings (which we term BioFLAIR[2]) on a variety of bio NER tasks and compare those with results from BERT-type networks. We also investigate the effects of a small amount of additional pretraining on PubMed content, and of combining FLAIR and ELMO models. We find that with the provided embeddings, FLAIR performs on-par with the BERT networks – even establishing a new state of the art on one benchmark. Additional pretraining did not provide a clear benefit, although this might change with even more pretraining being done. Stacking the FLAIR embeddings with others typically does provide a boost in the benchmark results.

## 1. Introduction

There continues to be a rapid increase in the number of biomedical publications each year, making NLP an essential tool for large-scale knowledge extraction and machine reading of this scientific literature. Deep learning has gained popularity among researchers as it has shown to boost the development of effective biomedical text mining models. But training these deep learning models often requires large amounts of labeled data, which is very hard to acquire in scientific fields like Biology and Medicine. As an alternative to the supervised learning approach, ELMo (2), GPT (3) and BERT (4) have shown the effectiveness of semi-supervision. In that approach, a large corpus of unlabeled text is used to pretrain a model. That model can then be finetuned to the specifics of an individual benchmark task such as Gene or Protein detection, POS tagging, etc. This has been shown to significantly improve performance on many NLP tasks (14).

FLAIR (1) is an alternative unsupervised embedding method which has not been as well studied as the others. FLAIR has seen to achieve state-of-the-art results for many common NER datasets like Conll-03(8), Ontonotes, etc., but we have not seen results for FLAIR on biomedical NER

---

benchmarks. In this paper we evaluate the most recent version of FLAIR on several bio NER benchmarks to compare its performance with that of BERT and other well-known methods.

## 2. Method

We downloaded the most recent version of FLAIR (0.4.2) from its GitHub site, https://github.com/zalandoresearch/flair. The FLAIR authors make a number of pretrained models available. We initially use the 'pubmed-x' embeddings which they provide. These were trained on about 5% of the PubMed abstracts from before 2015. We refer to this combination as BioFLAIR v1. Having a model pretrained on biomedical material is very important, as it has been shown (14, 15) that models pretrained on the same kind of material as will be used in the fine-tuning makes a significant difference in performance.

The pubmed-x embeddings are based on much less training than the BioBERT (14) embeddings. Time and budget did not allow for an extensive pre-training, but we did look at the effect of some additional pretraining. Table 1 lists the three corpora extracted from PubMed. The first is the corpus used for the pubmed-x embeddings.

| Name Of Corpus | Details |
| --- | --- |
| PubMed (V1) | 1,219,734 PubMed abstracts: 5% sample of PubMed abstracts until 2015 |
| PubMed (V2) | 261,736 PubMed abstracts: sample of PubMed abstracts after 2000 |
| PubMed (V3) | 281,394 PubMed abstracts: sample of PubMed abstracts after 2000 |

Table 1: Size of pretraining corpora

BioBERT (14) and SciBERT (15) tested on several benchmarks, including a number of NER tasks. We selected six of these benchmarks for testing with FLAIR in order to make comparison as simple as possible. The benchmark corpora used are listed below in Table 2. The benchmark corpora were downloaded. Their training splits were used to fine-tune separate models for each benchmark, which were then evaluated on each test split. These benchmark corpora are not large. We also tested the effect of combining similar corpora (NCBI-Disease and BC5DR-Disease) to obtain a larger test corpus.

| Dataset | Entity Type | # sents |
| --- | --- | --- |
| NCBI-disease (9) | Disease | 7,287 |
| BC5CDR-disease (10) | Disease | 15,030 |
| BC5CDR (10) | Disease and Drug/Chemical | 15,030 |
| JNLPBA (11) | Gene/Protein | 24,806 |
| Species-800 (12) | Species | 8,193 |
| LINNAEUS (13) | Species | 23,152 |

Table 2: The entity types and size of each benchmark corpus used for evaluation. The sentence counts are totals, including the train, dev and test splits.

The NCBI-disease corpus is the smallest of those used by BioBERT. We used that corpus to test a few configurations of FLAIR and its pretraining corpus. Based on those results a single configuration was selected, then that configuration was used for the other benchmarks of Table 2. The configurations tested are shown in Table 3.

| Model Name | Model Description |
|---|---|
| BioFLAIR (V1) [3] | FLAIR's pubmed-x model, pretrained on PubMed (V1) |
| BioFLAIR (V2) | BioFLAIR (V1) further trained on PubMed (V2) |
| BioFLAIR (V3) | BioFLAIR (V2) further trained on PubMed (V3) |
| CRAWL | FastText embeddings over Web crawls |
| BioELMo [4] | PubMed model for ELMo |
| MULTI | FLAIR embeddings trained on a mix of corpora (Web, Wikipedia, Subtitles, News) |

Table 3: Models tested on NCBI-Disease Corpus

## 3. Results

The results of our first experiment, on adding additional pretraining data, are shown below in the first three rows of Table 4. FLAIR's provided pubmed-x model is indicated by BioFLAIR (V1). This outscored the models with additional PubMed pretraining. Note, however, that the score does not decline monotonically. This may indicate that additional pretraining would eventually be of benefit.

The last three rows of Table 4 show the effect of stacking different embeddings. Here we uniformly see a benefit to combining some other kind of embeddings. The ELMo embeddings seem particularly beneficial. It should be noted however, that this requires significantly more computation than the FLAIR embeddings alone.

| Effect Of | Dataset | Model Details | F1 score |
|---|---|---|---|
| Additional Pretraining data | NCBI | BioFLAIR (V1) | **87.33** |
| | NCBI | BioFLAIR (V2) | 86.99 |
| | NCBI | BioFLAIR (V3) | 87 |
| Different Stacked Embeddings | NCBI | BioFLAIR (V1) + CRAWL | 88.18 |
| | NCBI | BioFLAIR (V3) + BioELMo | 88.3 |
| | NCBI | BioFLAIR (V1) + BioELMo | **88.47** |

Table 4: Studying effect of amount of pretraining data and different embeddings on NCBI disease NER dataset.

The results of our main experiment – comparing FLAIR with BERT – are shown below in Table 5. We have multiple BERT variants. Two are from the BioBERT paper (14) and one from the SciBERT paper (15). These models are the state of the art on these benchmarks at this time. We compare with the BioFLAIR model that had the best performance on the NCBI-Disease benchmark. That used the provided pubmed-x embeddings, stacked with BioELMo embeddings). The best-scoring configuration for each benchmark is shown in bold text, the second best is underlined.

---

[3] https://github.com/zalandoresearch/flair/pull/519
[4] https://github.com/zalandoresearch/flair/pull/503

We see that the FLAIR+ELMo model is very competitive with the BERT models. It typically trails only the BioBERT model that was trained on much more material, and it outscores all BERT variants on the BC5DR and Species-800 benchmarks. We even set a new state of the art score for the Species-800 benchmark.

| Benchmark | BioBERT (PubMed+ PMC) | BioBERT (PubMed) | SciBERT | MULTI | Ours (BioFLAIR (V1) + BioELMo) |
|---|---|---|---|---|---|
| NCBI | **89.36** | 87.38 | 86.91 | 84.72 | 88.85[5] |
| BC5CDR-disease | **86.56** | 86.20 | NA | NA | 85.31 |
| BC5CDR | NA | NA | 88.94 | 82.93 | **89.42** |
| JNLPBA | **77.59** | 76.65 | 75.95 | 74.14 | 77.03 |
| Species-800 | 75.31 | 73.08 | NA | 75.7 | **82.44** |
| LINNAEUS | **89.81** | 88.13 | NA | NA | 87.02 |

Table 5: Test results in biomedical named entity recognition. We compared our results with BioBERT (14) and SciBERT (15), which are similar in nature and have current state-of-the-art scores for these datasets.

The results of our final experiment, combining similar benchmark corpora to obtain larger training and test sets, are shown in Table 6. We do not see a consistent benefit from this. Adding the BC5DR-Disease training material to the NCBI-Disease training material does boost the score on the NCBI-Disease test. However, the converse is not true. In fact, the decline in the BC5DR-Disease scores after merging is greater than the improvement in the NCBI-Disease score.

| Dataset | Model Details | F1 score |
|---|---|---|
| NCBI | BioFLAIR (V1) + BioELMo | 88.47 |
| NCBI (+BC5DR-disease) | BioFLAIR (V1) + BioELMo | **88.85** |
| BC5DR-disease | BioFLAIR (V1) + BioELMo | **85.31** |
| BC5DR-disease (+ NCBI) | BioFLAIR (V1) + BioELMo | 84.49 |

Table 6: Result after adding more training data by combing similar datasets.

## 4. Discussion

Our main result is that the FLAIR embeddings are competitive with those from BERT. In an ensemble with ELMo, they occasionally even gave better results. It is not surprising that the combination outperforms the single FLAIR model, indeed it would be surprising if that were not the case. What is surprising is how close the results were, considering how little pretraining was performed on the FLAIR model compared to that of BioBERT.

We saw that increasing the amount of pretraining data didn't automatically improve the scores on the downstream benchmarks. The relation was more complex, with some additional data harming results and then still more data starting to recover those losses. More analysis, including hyperparameter optimization, is needed to understand and explain this behavior.

Not surprisingly, we saw that combining the training material from different benchmark tasks, even ones as similar as the disease identification benchmarks, did not automatically improve the scores. Here again, more analysis of the benchmarks and the hyperparameters is needed to fully understand this. Intuitively, the difficulties in multi-task training make it unsurprising that a simple combination of data would not yield improvements.

---
[5] Trained on NCBI (+BC5DR-disease) dataset.

## 5. Conclusion and Future Work

In this paper, we introduce BioFLAIR, a pretrained pooled contextualized embedding model for Biomedical Sequence Labeling tasks like NER. It achieves near state-of-the-art scores, with a much lighter model than the current SOTAs.

We expect we could improve performance with more pretraining; in line with that of BioBERT. There are other lines of investigation as well. Applying the FLAIR embeddings to tasks other than NER, such as relation classification and question answering, would be interesting. Comparison against more recent models, such as XLNet, is another line of investigation. Ensembles of all combinations of BERT, XLNet, ELMo, FastText, FLAIR, etc. would be interesting to see which ones are more or less like the others and bring different value to the ensemble. It would also be interesting to see how well it works when applied to other scientific domains.

## References

[1] Akbik, Alan, et al. "FLAIR: An Easy-to-Use Framework for State-of-the-Art NLP." *Proceedings of the 2019 Conference of the North American Chapter of the Association for Computational Linguistics (Demonstrations)*. 2019.

[2] Peters, Matthew E., et al. "Deep contextualized word representations." *arXiv preprint arXiv:1802.05365* (2018).

[3] Radford, Alec, et al. *Improving language understanding with unsupervised learning*. Technical report, OpenAI, 2018.

[4] Devlin, Jacob, et al. "Bert: Pre-training of deep bidirectional transformers for language understanding." *arXiv preprint arXiv:1810.04805* (2018).

[5] Mikolov, Tomas, et al. "Efficient estimation of word representations in vector space." *arXiv preprint arXiv:1301.3781* (2013).

[6] Pennington, Jeffrey, Richard Socher, and Christopher Manning. "Glove: Global vectors for word representation." *Proceedings of the 2014 conference on empirical methods in natural language processing (EMNLP)*. 2014.

[7] Akbik, Alan, Tanja Bergmann, and Roland Vollgraf. "Pooled contextualized embeddings for named entity recognition." *Proceedings of the 2019 Conference of the North American Chapter of the Association for Computational Linguistics: Human Language Technologies, Volume 1 (Long and Short Papers)*. 2019.

[8] Sang, Erik F., and Fien De Meulder. "Introduction to the CoNLL-2003 shared task: Language-independent named entity recognition." *arXiv preprint cs/0306050* (2003).

[9] Doğan, Rezarta Islamaj, Robert Leaman, and Zhiyong Lu. "NCBI disease corpus: a resource for disease name recognition and concept normalization." *Journal of biomedical informatics* 47 (2014): 1-10

[10] Li, Jiao, et al. "BioCreative V CDR task corpus: a resource for chemical disease relation extraction." *Database* 2016 (2016).

[11] Kim, Jin-Dong, et al. "Introduction to the bio-entity recognition task at JNLPBA." *Proceedings of the international joint workshop on natural language processing in biomedicine and its applications*. Association for Computational Linguistics, 2004.

[12]     Pafilis, Evangelos, et al. "The SPECIES and ORGANISMS resources for fast and accurate identification of taxonomic names in text." *PLoS One* 8.6 (2013): e65390.

[13]     Gerner, Martin, Goran Nenadic, and Casey M. Bergman. "LINNAEUS: a species name identification system for biomedical literature." *BMC bioinformatics* 11.1 (2010): 85.

[14]     Lee, Jinhyuk, et al. "Biobert: pre-trained biomedical language representation model for biomedical text mining." *arXiv preprint arXiv:1901.08746* (2019).

[15]     Beltagy, Iz, Arman Cohan, and Kyle Lo. "Scibert: Pretrained contextualized embeddings for scientific text." *arXiv preprint arXiv:1903.10676* (2019).

.